\documentclass[runningheads]{llncs}

\usepackage{eccv}

\usepackage{eccvabbrv}

\usepackage{graphicx}
\usepackage{booktabs}

\usepackage[accsupp]{axessibility}  

\usepackage{hyperref}

\usepackage{orcidlink}
\graphicspath{{./figures/}{./supp_figs/}{./rebt_figs/}}

\usepackage[group-separator={,}]{siunitx}
\usepackage{amsmath}
\usepackage{amsfonts}
\usepackage{makecell}
\usepackage{lipsum}
\usepackage{tcolorbox}
\usepackage{multirow}
\usepackage{marvosym}

\usepackage[capitalize]{cleveref}
\crefname{section}{Sec.}{Secs.}
\Crefname{section}{Section}{Sections}
\Crefname{table}{Table}{Tables}
\crefname{table}{Tab.}{Tabs.}
\Crefname{figure}{Figure}{Figures}
\crefname{figure}{Fig.}{Figs.}
\Crefname{equation}{Equation}{Equations}
\crefname{equation}{Eq.}{Eqs.}
\Crefname{algorithm}{Algorithm}{Algorithms}
\crefname{algorithm}{Alg.}{Algs.}

\usepackage[labelformat=simple]{subcaption}

\colorlet{lightpink}{pink!35}
\colorlet{lightcyan}{cyan!20}
\colorlet{lightgray}{gray!40}
\definecolor{darkgray}{rgb}{0.9, 0.9, 0.9}
\definecolor{lightgreen}{rgb}{0.886, 0.941, 0.851}
\colorlet{red}{red!80}
\colorlet{blue}{blue!80}
\colorlet{green}{green!60!black}
\colorlet{algemp}{cyan!10}
\colorlet{lightred}{red!50}

\newcommand{\msmall}[1]{{\fontsize{8pt}{10pt}\selectfont #1}}

\usepackage{dirtree}
\usepackage{xcolor,colortbl}
\usepackage{pifont}
\usepackage{wrapfig}

\newcolumntype{a}{>{\columncolor{gray!20!white}}c}

\begin{document}

\title{ProxyCLIP: Proxy Attention Improves CLIP \\
for Open-Vocabulary Segmentation} 

\titlerunning{ProxyCLIP}

\author{Mengcheng Lan\inst{1} \and
Chaofeng Chen\inst{1} \and
Yiping Ke\inst{2} \and 
Xinjiang Wang \inst{3} \and \\
Litong Feng \inst{3} \and
Wayne Zhang \inst{3,4}\thanks{Corresponding author.}
}

\authorrunning{M.Lan et al.}

\institute{S-Lab, Nanyang Technological University \and
CCDS, Nanyang Technological University \ \ \ \ 
\inst{3} SenseTime Research \\
\inst{4} Guangdong Provincial Key Laboratory of Digital Grid Technology \\
\email{lanm0002@e.ntu.edu.sg}\ \  
\email{\{chaofeng.chen,\ ypke\}@ntu.edu.sg}\\
\email{\{wangxinjiang, fenglitong, wayne.zhang\}@sensetime.com}
\url{https://github.com/mc-lan/ProxyCLIP}
}

\maketitle

\begin{abstract}
    Open-vocabulary semantic segmentation requires models to effectively integrate visual representations with open-vocabulary semantic labels.
    While Contrastive Language-Image Pre-training (CLIP) models shine in recognizing visual concepts from text, they often struggle with segment coherence due to their limited localization ability. 
    In contrast, Vision Foundation Models (VFMs) excel at acquiring spatially consistent local visual representations, yet they fall short in semantic understanding.
    This paper introduces ProxyCLIP, an innovative framework designed to harmonize the strengths of both CLIP and VFMs, facilitating enhanced open-vocabulary semantic segmentation. 
    ProxyCLIP leverages the spatial feature correspondence from VFMs as a form of proxy attention to augment CLIP, thereby inheriting the VFMs' robust local consistency and maintaining CLIP's exceptional zero-shot transfer capacity.
    We propose an adaptive normalization and masking strategy to get the proxy attention from VFMs, allowing for adaptation across different VFMs.
    Remarkably, as a \emph{training-free} approach, ProxyCLIP significantly improves the average mean Intersection over Union (mIoU) across eight benchmarks \textbf{from 40.3 to 44.4}, showcasing its exceptional efficacy in bridging the gap between spatial precision and semantic richness for the open-vocabulary segmentation task.
  \keywords{Semantic segmentation \and Vision language model \and Vision foundation model \and Open vocabulary}
\end{abstract}

\section{Introduction}
\label{sec:intro}
Open-vocabulary semantic segmentation \cite{zhou2022extract,xu2022groupvit} aims to partition an image into coherent segments and assign each segment a semantic label from arbitrary visual concepts. 
The success of this task hinges on a model's ability to generate semantically coherent visual representations that align well with textual descriptions.
The emergence of Contrastive Language-Image Pre-training (CLIP) models \cite{jia2021scaling,radford2021learning, cherti2023reproducible, xu2023demystifying}, trained on large-scale image-text pair datasets, has showcased remarkable capabilities in semantic understanding, such as image recognition \cite{radford2021learning} and image-text retrieval \cite{cho2021unifying, li2022blip}.
However, while excelling in recognizing concepts within images, CLIP models often struggle with producing coherent image segments for dense prediction tasks due to limitations in localization ability.
In contrast, Vision Foundation Models (VFMs) like self-supervised methods \cite{caron2021emerging, oquab2023dinov2,he2022masked} and the Segment Anything Model (SAM) \cite{kirillov2023segment} have demonstrated proficiency in learning spatially consistent visual representations.
These models have found extensive applications in unsupervised semantic segmentation \cite{hamilton2022unsupervised,lan2024smooseg} and object detection \cite{simeoni2021localizing,wang2023tokencut,simeoni2023unsupervised} to generate \textit{class-agnostic} image segments.
However, they often lack semantic understanding, requiring fine-tuning on downstream tasks.

Significant efforts have been dedicated to enhance the open-vocabulary semantic segmentation capabilities of CLIP models. 
A key research area focuses on exploiting the inherent localization properties of CLIP. 
This involves modifying its attention mechanism during the inference stage to rearrange local spatial information \cite{zhou2022extract,wang2023sclip,bousselham2023grounding,li2023clip}, or fine-tuning the CLIP model to improve its local consistency via the self-distillation technique \cite{chen2023exploring,mukhoti2023open,wu2023clipself,naeem2023silc}.
Despite their efficacy, these methods are constrained by the intrinsic performance of CLIP models, as the contrastive learning objective does not explicitly encourage learning spatially consistent local features \cite{caron2021emerging}.
Another research path emphasizes improving CLIP representations with semantically coherent properties through knowledge distillation \cite{wang2023samclip,wysoczanska2024clip,yuan2024open} from VFM features \cite{caron2021emerging,kirillov2023segment}. 
While these fine-tuning approaches are usually more effective, they carry the potential risk of sacrificing the strength of the original models during fine-tuning, which may yield sub-optimal results.

In this study, our aim is to integrate the expertise of CLIP models and VFMs in \emph{a training-free manner} without compromising their individual strengths for the open-vocabulary semantic segmentation. 
The key principle is to borrow the robust feature correspondence properties from VFMs and maintain the zero-shot transfer ability of CLIP. 
We first verify the feasibility of our idea by comparing the semantic coherence between CLIP's attention scores and VFMs' feature correspondence maps.
Our investigation reveals that VFMs' feature correspondence significantly surpasses various types of attention scores in CLIP, highlighting the significant potential of their integration. 
Consequently, we propose ProxyCLIP, a general framework designed to harness VFMs' feature correspondence as a proxy attention for CLIP.

Designing a universal proxy attention from various VFMs is challenging due to their diverse inductive biases in visual representations, which leads to the difficulties in ensuring the consistency and separateness of the proxy attention.
Our research explores different configurations, leading to two crucial designs for improved performance in the open-vocabulary segmentation:
\begin{itemize}
    \item[\textbullet] We introduce the adaptive normalization and masking strategy to the VFMs' feature correspondence, ensuring our proxy attention remains effective and consistent across different VFMs.
    \item[\textbullet] We find that VFM architectures with smaller patch size produce clearer boundaries, which contributes to superior segmentation results.
\end{itemize}
To our knowledge, ProxyCLIP is the first training-free method to successfully achieve better performance with larger CLIP models, while previous approaches such as MaskCLIP \cite{zhou2022extract} fail to benefit from stronger CLIP backbones. 
Extensive experiments conducted across 8 segmentation benchmarks establish ProxyCLIP as a new state-of-the-art solution for the open-vocabulary segmentation.

\section{Related Work}
\label{sec:relatedwork}
\subsubsection{Vision-language pre-training models.}
Vision-Language Models (VLMs) constitute a significant stride in AI, designed to imbibe generic multimodal representations through extensive training on large-scale image-text pair datasets. 
Recent advancements in contrastive vision-language pre-training, facilitated by dual encoders, have significantly enhanced the generalization capabilities of image recognition systems \cite{jia2021scaling, li2021align, yao2021filip, yuan2021florence, radford2021learning, zhai2022lit, cherti2023reproducible, xu2023demystifying}.
Among these advancements, the CLIP series \cite{radford2021learning, cherti2023reproducible, xu2023demystifying} stands out. 
Pre-trained on billion-scale image-text pairs, CLIP models exhibit exceptional performance in zero-shot image classification and image-text retrieval tasks. 
However, despite their success, the reliance on image-level supervision often poses challenges in aligning local image regions with corresponding text, leading to subpar performance in dense prediction tasks \cite{zhou2022extract, zhong2022regionclip, xu2023open}.
Towards this challenge, various weakly-supervised approaches have emerged, aiming to indirectly align region and language representations using solely image-text pairs \cite{mukhoti2023open, wu2023clipself}. 
Additionally, other works \cite{liu2023grounding, zhang2022glipv2, rasheed2023glamm} have sought to perform region-text alignment leveraging visual grounding datasets\cite{plummer2015flickr30k,krishna2017visual}. 

\subsubsection{Vision foundation models.}
The field of computer vision has witnessed the rise of potent models built upon Vision Transformers (ViT) \cite{dosovitskiy2020image}.
Recent advancements in self-supervised learning (SSL) \cite{caron2021emerging, oquab2023dinov2, darcet2023vision, zhou2021ibot, he2022masked} have yielded features with strong localization properties, opening avenues for novel applications across various downstream tasks.
These features have been effectively harnessed in diverse areas, including unsupervised semantic segmentation \cite{hamilton2022unsupervised, yin2022transfgu, lan2024smooseg}, unsupervised object detection \cite{simeoni2021localizing, wang2023tokencut, simeoni2023unsupervised}, unsupervised instance segmentation \cite{wang2023cut, wang2022freesolo}, and zero-shot semantic segmentation \cite{wysoczanska2024clip, wysoczanska2023clip, wang2024image, barsellotti2024training}.
DINO \cite{caron2021emerging} and DINOv2 \cite{oquab2023dinov2} have contributed significantly to the advancement of SSL models, demonstrating remarkable capabilities in producing spatially coherent visual representations. 
More recently, the Segment Anything Model (SAM) \cite{kirillov2023segment} has emerged as proficient in the object segmentation, adeptly producing precise masks within images, albeit limited to class-agnostic masks.
Our work introduces a generic framework designed to harness the spatial coherence properties of VFMs' features to enhance the dense prediction performance of CLIP models.

\subsubsection{Open-vocabulary semantic segmentation.}
The emergence of vision-language pre-training models has spurred recent advancements in open-vocabulary semantic segmentation methods.
Many studies \cite{zhou2022extract,li2023clip, wang2023sclip, bousselham2023grounding,li2024cascade,sun2024clip} aim to leverage CLIP's inherent localization properties for the dense prediction tasks. 
For instance, MaskCLIP \cite{zhou2022extract} demonstrates that the \textit{value} embeddings of the last layer of CLIP exhibit superior localization capabilities compared to token embeddings, resulting in impressive performance in open-vocabulary dense prediction tasks.
Additionally, some researchers \cite{li2023clip, wang2023sclip, bousselham2023grounding} have identified noisy attention maps as a bottleneck in CLIP's performance on dense prediction.
To address this, they propose to revise the vanilla \textit{query}-\textit{key} attention to \textit{self}-\textit{self} attention, such as \textit{value}-\textit{value} \cite{li2023clip}, \textit{query}-\textit{query}, \textit{key}-\textit{key} attention, or their combination \cite{wang2023sclip, bousselham2023grounding}, leading to substantial performance improvements.
Despite these efforts, 
these methods are constrained by the intrinsic performance of CLIP models.
Besides these training-free methods, various weakly-supervised retraining methods \cite{mukhoti2023open, cha2023learning, ren2023viewco, xu2022groupvit, luo2023segclip, xu2023learning, zhang2023uncovering} have been proposed to address the limitations of CLIP.
These methods incorporate group tokens \cite{xu2022groupvit, luo2023segclip, xu2023learning} or prototype tokens \cite{zhang2023uncovering} into the image encoder to enhance object discovery and image segmentation. 
However, this enhancement may come at the cost of compromising some of CLIP's open-vocabulary properties.
Related approaches include recent CLIP-DINOiser \cite{wysoczanska2024clip} and SAM-CLIP \cite{wang2023samclip}, which integrate the favorable spatial consistent properties of DINO \cite{caron2021emerging} and SAM \cite{kirillov2023segment} with CLIP through knowledge distillation techniques, aiming to enhance the local patch representations of images for the dense prediction tasks.
Besides, ReCo \cite{shin2022reco} proposes to dynamically curate training sets by retrieving unlabeled images using CLIP, and to co-segment all entities among these collections based on the visual representations of DINO \cite{caron2021emerging}.
In this work, we attempt to seamlessly integrate the strengths of CLIP and VFMs in a training-free manner by designing a novel proxy attention module, without compromising their unique strengths, ultimately striving to achieve state-of-the-art results.

\section{Method}
\label{sec:method}

\subsection{Preliminaries and Motivation}
\label{sec:preliminary}
\subsubsection{CLIP's representations.}
\label{sec:vit}
The CLIP ViT architecture primarily consists of residual attention blocks.
Let $\boldsymbol{x} = [x_0, x_1, \dots, x_L]^T$ denotes the input to the last block, where $x_0$ represents the global class embedding, and $\{x_i| i= 1, 2, \dots, L\}$ denotes local patch embeddings.
The forward process of a residual attention block is as follows:
\begin{align}
    &\boldsymbol{q} = \textup{Emb}_q(\boldsymbol{x}),\ \boldsymbol{k} =\textup{Emb}_k(\boldsymbol{x}),\ \boldsymbol{v} = \textup{Emb}_v(\boldsymbol{x}), \\
    &\boldsymbol{y} = \boldsymbol{x} + \textup{Proj}(\textup{Attn}_{qk}\cdot \boldsymbol{v}), \\
    &\boldsymbol{z} = \boldsymbol{y} + \textup{FFN}(\textup{LN}(\boldsymbol{y})),
\end{align}
where $\boldsymbol{q}$, $\boldsymbol{k}$ and $\boldsymbol{v}$ represent the \textit{query}, \textit{key} and \textit{value} embeddings, respectively.
$\textup{Emb}_\cdot$ comprises a layer norm (LN) and a projection layer, and FFN is a feed-forward network. 
$\textup{Attn}_{qk} = \textup{SoftMax}(\frac{\boldsymbol{q}\boldsymbol{k}^T}{\tau})$ denotes the attention score with a constant scaling factor $\tau$.
Typically, $z_{\textup{image}}=\boldsymbol{z}[0] \in \mathbb{R}^{1 \times d}$ is selected as the class representation with dimension $d$, representing the entire image, while $\boldsymbol{z}_{\textup{dense}} = \boldsymbol{z}[1:L] \in \mathbb{R}^{L \times d}$ denotes the feature representations.
For brevity, we omit the final output layer following the last attention block.
Additionally, the class representation is removed from the visual representations of both CLIP models and VFMs in the subsequent analysis.

\subsubsection{Discussion and motivation of ProxyCLIP.}
The image-level contrastive learning paradigm adopted by CLIP often encounters challenges in aligning local image patches with corresponding textual representations \cite{mukhoti2023open, zhou2022extract}. 
This difficulty arises from its global attention mechanism's indiscriminate propagation of information across all image patches, while effective at capturing the global context, tends to weaken the specificity of local patch representations. 
Recent studies \cite{zhou2022extract, li2023clip, bousselham2023grounding, wang2023sclip} suggest that modifying the attention mechanism to better organize the spatial information of local patches can benefit the open-vocabulary segmentation.
We hypothesize that an improved local patch representation should selectively incorporate contextually relevant patches sharing the same semantics, thus enhancing the precision of patch representations for dense prediction tasks.
However, CLIP's local feature correspondence is not ideal, which cannot ensure spatially coherent segments using CLIP alone. 
Meanwhile, recent vision foundation models like DINO \cite{caron2021emerging}, DINOv2 \cite{oquab2023dinov2, darcet2023vision}, and SAM \cite{kirillov2023segment} have demonstrated the ability to learn semantically coherent representations with strong localization capabilities. 
This motivates us to improve CLIP with these VFMs.

We begin with a comparative analysis of the semantic coherence within attention scores \cite{li2023clip, bousselham2023grounding, wang2023sclip} of CLIP and feature correspondences \cite{hamilton2022unsupervised} of VFMs, using data from the COCO stuff \cite{caesar2018coco} dataset.
Specifically, we collect all different attention scores at the last attention layer of CLIP, and extract all patch representations of VFMs for each image in COCO Stuff.
For CLIP, let $\textup{Attn}_{qk} \in \mathbb{R}^{L\times L}$,
\begin{wrapfigure}{r}{0.35\textwidth}
    \centering
    \includegraphics[width=0.35\textwidth]{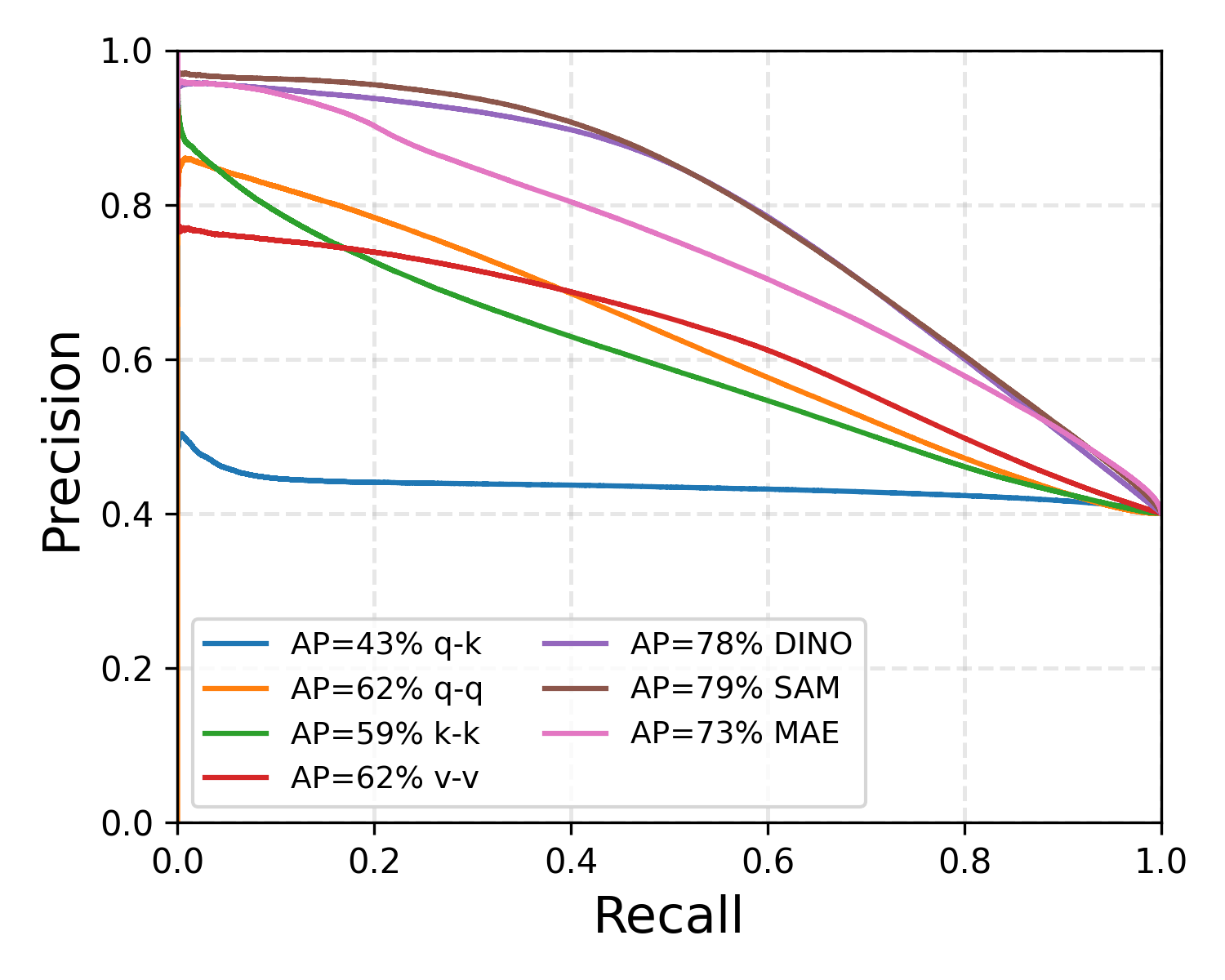}
    \caption{Precision recall curves of different classifiers. Higher average precision (AP) indicates better semantic correspondence.}
    \label{fig:pr_curve}
\end{wrapfigure}
as well as the self-self attention $\textup{Attn}_{qq}$, $\textup{Attn}_{kk}$ and $\textup{Attn}_{vv}$, signify the similarities between image patches.
For VFMs, we extract the patch representations 
$F \in \mathbb{R}^{L_v\times D_v}$ using their frozen backbones, where $L_v$ and $D_v$ denote the number and dimensionality of image patches, respectively.
We then compute their
cosine similarity $S_{ij} = \frac{F_i}{|F_i|} \frac{F_j}{|F_j|}$.
To evaluate the semantic coherence, we use the attention scores of CLIP and similarity scores of VFMs as a binary classifier to predict whether two patches share the same semantic label. 
We determine the ground true label for each patch through majority voting using the segmentation maps.
Given a pair of semantic labels $l_i$ and $l_j$, we assign a target value of 1 for the classifier if the labels $l_i = l_j$; otherwise, the target value is 0.

\begin{figure}[t]
    \centering
    \includegraphics[width=0.8\linewidth]{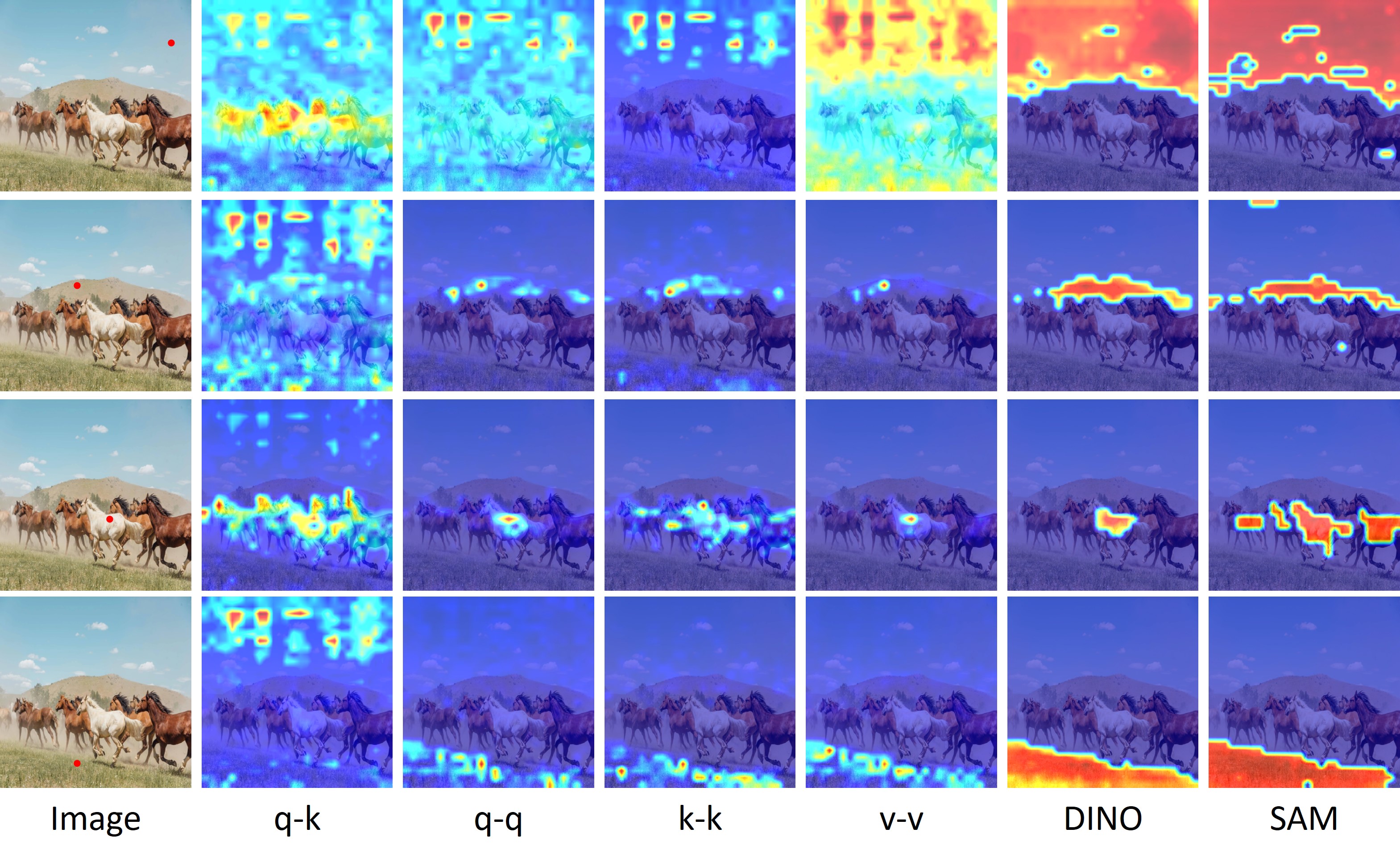}
    \caption{Attention scores (maps) between CLIP, DINO and SAM using different seeds (in \textcolor{red}{red}). For CLIP's attention maps, we display only the first head of multi-head self-attention maps.}
    \label{fig:attn}
\end{figure}

\cref{fig:pr_curve} presents the precision-recall curves for different attention scores of CLIP and the feature correspondence of VFMs using DINO \cite{caron2021emerging}, SAM \cite{kirillov2023segment}, and MAE \cite{he2022masked} backbones.
Notably, the self-self attention scores exhibit significantly higher average precision than the vanilla attention scores, reflecting their better semantic coherence, which focuses on contextually relevant information.
This finding aligns with earlier studies \cite{li2023clip, bousselham2023grounding, wang2023sclip} that reported improved outcomes over the vanilla CLIP. 
Remarkably, the feature correspondence of VFMs, especially with DINO and SAM backbones, far surpasses the attention scores of CLIP. 
We also showcase qualitative results in \cref{fig:attn}, illustrating that attention maps generated by DINO and SAM effectively highlight semantically relevant patches in different contexts, such as sky, mountain, horses, and grass. 
This is highly encouraging, as it suggests that the quality of CLIP’s dense patch representations could be enhanced by employing VFMs' feature representations as a proxy attention for CLIP.

\subsection{ProxyCLIP}
In this section, we detail our method for improving CLIP's dense representations by utilizing pre-trained Vision Foundation Models (VFMs).
Our objective is to seamlessly combine the advanced spatial coherence of VFMs with CLIP's semantic understanding capabilities within a training-free framework, thereby enhancing dense vision-language inference.

\subsubsection{Architecture.}
\begin{figure}[t]
    \centering
    \includegraphics[width=0.9\linewidth]{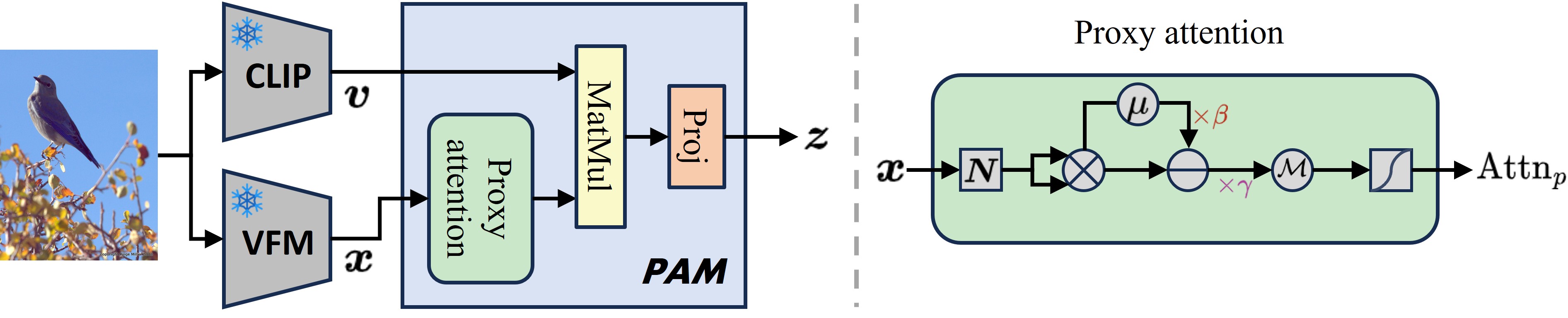}
    \caption{Overview of the ProxyCLIP architecture. ProxyCLIP consists of two frozen image encoders and a novel proxy attention module (PAM). On the right, the flow of the proxy attention mechanism with an adaptive normalization and masking strategy is illustrated, corresponding to \cref{eq:norm,eq:masking,eq:proxy_attention}.}
    \label{fig:architecture}
\end{figure}

The overall framework, illustrated in \cref{fig:architecture}, includes two frozen image encoders, \ie, CLIP image encoder and VFM backbone, and a novel proxy attention module (PAM). 
In a word, we expect CLIP's image encoder to provide the semantic base (\textit{value} embeddings), while the VFMs' image encoder contributes attention scores (\textit{query} and \textit{key} embeddings) to reorganize this semantic base. 
The goal is to produce semantically coherent local patch representations that align well with language cues.

\subsubsection{Proxy attention module.}
The Proxy Attention Module (PAM) serves as a pivotal component in our approach.
PAM is designed to integrate visual representations $\boldsymbol{x} \in \mathbb{R}^{L_x\times D_x}$ from the VFM backbones and \textit{value} embeddings $\boldsymbol{v} \in \mathbb{R}^{n\times L_v\times D_v}$ from CLIP's last attention layer.
Here, $L_x$ and $L_v$ denote the length of the visual sequences, $D_x$ and $D_v$ represent their respective dimensions, and $n$ is the number of heads for $\boldsymbol{v}$.
The primary objective of PAM is to generate dense representations of images for subsequent vision-language inference tasks.
Let $\ell_2$-normalized visual representations $\boldsymbol{x}$ as the \textit{query} and \textit{key} embeddings,
the process of the proxy attention module is formulated as follows:
\begin{align}
    \textup{Attn}_{p} &= \textup{SoftMax}(\boldsymbol{x}\boldsymbol{x}^T),
    \label{eq:attn}\\
    \boldsymbol{z} &= \textup{Proj}(\textup{Attn}_{p}\cdot \boldsymbol{v}),
    \label{eq:z}
\end{align}
where $\textup{Attn}_p \in \mathbb{R}^{L_x\times L_x}$ represents the proxy attention score, which is expanded to encompass $n$ heads in a new axis.
Proj denotes a projection layer responsible for fusing the multiple self-attention outputs into the final representations,
\begin{wrapfigure}{r}{0.5\textwidth}
    \centering
  \begin{subfigure}{0.48\linewidth}
    \includegraphics[width=1.0\linewidth]{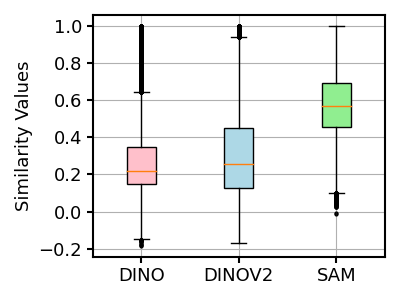}
    \label{fig:similarity_ori}
  \end{subfigure}
  \begin{subfigure}{0.48\linewidth}
  \includegraphics[width=1.0\linewidth]{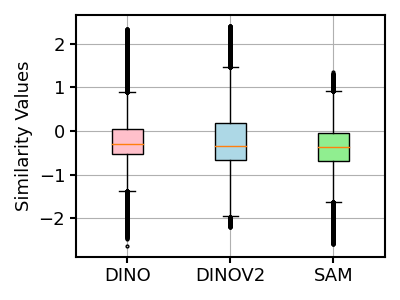}
    \label{fig:similarity_norm}
  \end{subfigure}
  \caption{The statistics of similarity matrix before (left) and after (right) normalization.}
  \label{fig:similarity}
\end{wrapfigure}
and its parameters are inherited from the output projection layer of CLIP's last self-attention layer.
It's noteworthy that our PAM differs from the cross-attention mechanism in existing works \cite{vaswani2017attention, kirillov2023segment,liu2023grounding}, which typically utilize a pair of keys and values from one encoder while employing the queries from another.

\subsubsection{Normalization and masking.}
In practice, we observed that the attention score calculated by \cref{eq:attn} may not always ensure good consistency and separateness across various VFMs due to their diverse inductive biases in visual representations.
For example, as illustrated in \cref{fig:similarity}, when analyzing a subset of the COCO Stuff dataset, we found that the median cosine similarity among patch representations learned from DINO with ViT-B/16 architecture is 0.22, whereas it is 0.57 for SAM with the same architecture.
To this end, we propose an adaptive normalization and masking approach to the similarity matrix:
\begin{align}
    A = \gamma(\boldsymbol{x}\boldsymbol{x}^T - \frac{\beta }{L_v^2}\sum_{i,j}[\boldsymbol{x}\boldsymbol{x}^T]_{ij}),
    \label{eq:norm}
\end{align}
\begin{equation}
    \mathcal{M}_{ij} = 
    \begin{cases}
        0, \ \ \ &A_{ij}\ge 0\\
        -\infty, \ \ \ &A_{ij}< 0
    \end{cases}
    \label{eq:masking}
\end{equation}
\begin{equation}
    \textup{Attn}_{p} = \textup{SoftMax}(A + \mathcal{M}).
    \label{eq:proxy_attention}
\end{equation}
Here, $\beta$ and $\gamma$ denote the shifting and scaling factors, respectively.
We illustrate the distribution of similarity matrices of different VFMs after normalization in \cref{fig:similarity}, showing their aligned statistics with similar median cosine similarity.
Based the normalized similarity matrix, we introduce the masking matrix $\mathcal{M}$ to suppress negative similarities in $A$. 
This ensures that each patch representation positively integrates contextually relevant information, leading to improved discriminativeness in dense patch representations. 
To ensure that the attention scores concentrate exclusively on contextually relevant patches, which constitute a minor fraction of the total image patches, we empirically set these two parameters to $\beta=1.2$ and $\gamma=3$ as default values. 
Moreover, through experiments on parameter sensitivity, we find that these settings are robust.

\subsubsection{Different resolution.}
Given the flexibility of our framework, ProxyCLIP can utilize various VFMs and CLIP models.
We observed that proxy attention with higher spatial resolution enhances the performance. 
Specifically, we propose to interpolate the spatial resolution of $\boldsymbol{v}$ to align with the size of $\boldsymbol{x}$, resulting in $L_v = L_x$.
This modification allows ProxyCLIP to benefit from VFMs with smaller patch sizes, such as the DINO ViT-B/8 architecture. 
Consequently, ProxyCLIP is able to produce more detailed segmentation maps, as demonstrated in \cref{fig:patch_size}, where the segmentation maps produced by ProxyCLIP with DINO ViT-B/8 as the proxy attention outperform that with DINO ViT-B/16.

\subsubsection{Open-vocabulary segmentation.} 
To perform dense inference with CLIP, $\boldsymbol{z}$ is first projected to the aligned vision-language space, denoted as $\boldsymbol{z}_v$. 
Then, we obtain the textual representations $\boldsymbol{z}_t\in \mathbb{R}^{C\times d}$ by feeding the text prompt ``\textsl{a photo of a \{label names\}.}'' with different label names to CLIP's text encoder, where $C$ denotes the number of classes, and $d$ represents the dimension of the shared latent vision-language space.
Next, we conduct patch-level classification by computing the cosine similarity between $\boldsymbol{z}_v$ and $\boldsymbol{z}_t$, and determine the label for each patch based on the most similar prompt.
Finally, we obtain the segmentation maps $\mathcal{S} \in \mathbb{R}^{h\times w\times 1}$ for the image by reshaping and up-sampling.

\section{Experiments}
\label{sec:experiment}

\subsection{Experiment Setups}
\subsubsection{Benchmark Datasets.}
We conduct a comprehensive evaluation on eight well-established datasets following \cite{cha2023learning, wysoczanska2023clip}. 
These datasets are divided into two main categories.
(\textit{i}) With background class: 
PASCAL VOC \cite{everingham2012pascal} (VOC),  PASCAL Context \cite{mottaghi2014role} (Context) and COCO Object \cite{caesar2018coco} (Object).
(\textit{ii}) Without background class:
PASCAL VOC20 \cite{everingham2012pascal} (VOC20), PASCAL Context59 \cite{mottaghi2014role} (Context59),  COCO Stuff \cite{caesar2018coco} (Stuff), Cityscapes \cite{cordts2016cityscapes} (City) and ADE20K \cite{zhou2019semantic} (ADE).

\subsubsection{Implementation \& metric.}
Our implementations are based on MMSegmentation \cite{contributors2020mmsegmentation}. 
For image pre-processing, we resize the images to accommodate varying dataset specifications: a shorter side of 336 pixels for PASCAL and COCO datasets and 448 pixels for Cityscapes and ADE20K datasets.
We adopt a sliding window strategy with a 336$\times$336 window and 112$\times$112 stride to benefit both CLIP models and VFMs.
For all datasets, we construct textual descriptions using the standard ImageNet prompts \cite{radford2021learning} along with their corresponding class names.
Our method undergoes direct evaluation on the validation set of all datasets without the need of retraining or fine-tuning.
We adopt the mean Intersection over Union (mIoU) metric to evaluate the semantic segmentation performance and report the results without any post-processing step.

\subsubsection{Comparison Methods.}
We compare our method against several baselines and state-of-the-arts. 
The baselines include CLIP \cite{radford2021learning} with ViT-B/16 and ViT-L/14 architectures, and OpenCLIP \cite{cherti2023reproducible} with the ViT-H/14 architecture.
Additionally,
We compare our method with two categories of state-of-the-art approaches.
\begin{itemize}
    \item \textbf{Training-free} methods: MaskCLIP \cite{zhou2022extract}, CLIPSurgery \cite{li2023clip}, GEM \cite{bousselham2023grounding}, ReCo \cite{shin2022reco} and SCLIP \cite{wang2023sclip};
    \item \textbf{Weakly-supervised training-based} methods: GroupViT \cite{xu2022groupvit}, 
    SegCLIP \cite{luo2023segclip}, ViewCo \cite{ren2023viewco}, OVSegmentor \cite{xu2023learning}, CoCu \cite{xing2023rewrite}, TCL \cite{cha2023learning}, SAM-CLIP \cite{wang2023samclip} and CLIP-DINOiser \cite{wysoczanska2023clip}.
\end{itemize}

For baseline methods and methods based on ViT-L/14 and ViT-H/14 architectures, we report their results based on our implementation. 
For state-of-the-art methods, 
if not otherwise specified, all reported results are directly sourced from the respective papers.

\subsection{Main results}

\begin{table}[!t]
  \caption{Open-vocabulary semantic segmentation comparison on 8 datasets. $^\dagger$ Results are directly cited from TCL \cite{cha2023learning}. 
  }
  \tabcolsep1.8pt
  \label{tab:main_results}
  \centering
  \begin{tabular}{l|ccccccccc}
    \toprule
    Methods & VOC & Context & Object & VOC20 & Context59 & Stuff & City & ADE & Avg. \\
    \midrule
    \rowcolor{lightgray} \multicolumn{2}{l}{Training-based}  &&&&&&&& \\
    SegCLIP \cite{luo2023segclip} & 52.6 & 24.7 & 26.5 & - & - & - & - & - & -\\
    ViewCo \cite{ren2023viewco} & 52.4 & 23.0 & 23.5 & - & - & - & - & - & - \\
    OVSegmentor \cite{xu2023learning} & 53.8 & 20.4 & 25.1 & - & - & - & - & 5.6 & - \\
    CoCu \cite{xing2023rewrite} & 51.4 & 23.6 & 22.7 & - & - & 22.1 & 15.2 & 12.3 & - \\
    SAM-CLIP \cite{wang2023samclip} & 60.6 & 29.2 & - & - & - & - & - & 17.1 & -  \\
    GroupViT$^\dagger$ \cite{xu2022groupvit} & 50.4 & 18.7 & 27.5 & 79.7 & 23.4 & 15.3 & 11.1 & 9.2 & 27.7\\
    TCL \cite{cha2023learning} & 51.2 & 24.3 & 30.4 & 77.5 & 30.3 & 19.6 & 23.1 & 14.9 & 33.1\\
    CLIP-DINOiser \cite{wysoczanska2023clip} & 62.2 & 32.4 & 35.0 & 80.2 & 35.9 & 24.6 & 31.7 & 20.0 & 40.3 \\
    \midrule
    \rowcolor{lightgray} \multicolumn{2}{l}{CLIP-ViT-B/16}  &&&&&&&& \\
    CLIPSurgery \cite{li2023clip} & - & 29.3 & - & - & - & 21.9 & 31.4 & - & - \\
    GEM \cite{bousselham2023grounding} & 46.2 & 32.6 & - & - & - & -& - & 15.7 & - \\
    ReCo$^\dagger$ \cite{shin2022reco} & 25.1 & 19.9 & 15.7 & 57.7 & 22.3 & 14.8 & 21.1 & 11.2 & 23.5\\
    CLIP\cite{radford2021learning} & 16.4 & 8.4 & 5.6 & 41.9 & 9.2 & 4.4 & 5.0 & 2.9 & 11.7 \\
    MaskCLIP$^\dagger$ \cite{zhou2022extract} & 38.8 & 23.6 & 20.6 & 74.9 & 26.4 & 16.4 & 12.6 & 9.8 & 27.9 \\
    SCLIP \cite{wang2023sclip} & 59.1 & 30.4 & 30.5 & 80.4 & 34.2 & 22.4 & 32.2 & 16.1 & 38.2 \\
    \rowcolor{lightgreen} ProxyCLIP & 61.3 & 35.3 & 37.5 & 80.3 & 39.1 & 26.5 & 38.1 & 20.2 & 42.3 \\
    \midrule
    \rowcolor{lightgray} \multicolumn{2}{l}{CLIP-ViT-L/14}  &&&&&&&& \\
    CLIP\cite{radford2021learning} & 8.2 & 4.1 & 2.7 & 15.6 & 4.4 & 2.4 & 2.5 & 1.7 & 5.2 \\
    MaskCLIP \cite{zhou2022extract} & 23.3 & 11.7 & 7.2 & 29.4 & 12.4 & 8.8 & 11.5 & 7.2 & 13.9 \\
    SCLIP \cite{wang2023sclip} & 43.5 & 22.3 & 25.0 & 69.1 & 25.2 & 17.6 & 18.6 & 10.9 & 29.0\\
    \rowcolor{lightgreen} ProxyCLIP & 60.6 & 34.5 & \textbf{39.2} & 83.2 & 37.7 & 25.6 & 40.1 & 22.6 & 43.0 \\
    \midrule
    \rowcolor{lightgray} \multicolumn{2}{l}{OpenCLIP-ViT-H/14} &&&&&&&& \\
    OpenCLIP \cite{cherti2023reproducible} & 8.8 & 5.0 & 5.3 & 21.7 & 5.5 & 3.2 & 4.3 & 2.8 & 7.1\\
    MaskCLIP \cite{zhou2022extract} & 31.4 & 13.3 & 16.2 & 41.7 & 15.8 & 8.4 & 17.7 & 10.4 & 19.3  \\
    SCLIP \cite{wang2023sclip} & 43.8 & 23.5 & 24.6 & 67.5 & 25.6 & 16.8 & 19.5 & 11.3 & 29.1 \\
    \rowcolor{lightgreen} ProxyCLIP & \textbf{65.0} & \textbf{35.4} & 38.6 & \textbf{83.3} & \textbf{39.6} & \textbf{26.8} & \textbf{42.0} & \textbf{24.2} & \textbf{44.4} \\
  \bottomrule
  \end{tabular}
\end{table}

\subsubsection{Quantitative results.}
We first conduct experiments to evaluate ProxyCLIP against various open-vocabulary semantic segmentation models.
The VFM adopted in our ProxyCLIP is DINO with ViT-B/8 architecture for its good performance.
\cref{tab:main_results} summarises the comparison results of all comparison methods on eight datasets, where our ProxyCLIP consistently achieves the best performance.
Our investigation yields interesting findings:
1) Among trained-based methods, CLIP-DINOiser \cite{wysoczanska2023clip} emerges as a standout performer, with the average mIoU (40.3) significantly higher than TCL \cite{cha2023learning} (33.1). 
CLIP-DINOiser's success can be attributed to the improved CLIP features distilled from DINO features, benefiting from the good spatial consistency of DINO.
This highlights the importance and benefits of integrating VFMs with CLIP over approaches like self-distillation used in TCL.
Meanwhile, the proposed ProxyCLIP outperforms CLIP-DINOiser by a significant margin, with average improvements of \textbf{+2.0 mIoU, +2.7 mIoU, and +4.1 mIoU} under CLIP-ViT-B/16, CLIP-ViT-L/14, and OpenCLIP-ViT-H/14 architectures, respectively. 
This underscores ProxyCLIP's superior efficiency in merging the strengths of VFMs and CLIP.
2) Notably, all training-based methods fall short when compared to ProxyCLIP, suggesting potential compromises to CLIP's original zero-shot transfer properties during fine-tuning. 
In contrast, as a training-free approach, ProxyCLIP seamlessly combines the spatial consistency of VFMs with the visual language alignment strengths of CLIP, without sacrificing their advantages.
3) Among training-free methods, ProxyCLIP significantly outperforms all competitors, achieving an average mIoU improvement of 4.1 with CLIP-ViT-B/16 architectures. This improvement expands to 14.0 mIoU and 15.3 mIoU with CLIP-ViT-L/14 and OpenCLIP-ViT-H/14, respectively. Notably, ProxyCLIP is the first training-free method to effectively utilize larger CLIP models to boost performance. 
In conclusion, these results clearly affirm ProxyCLIP's efficacy in open-vocabulary segmentation tasks. 

\begin{figure}[!t]
    \centering
    \includegraphics[width=0.9\linewidth]{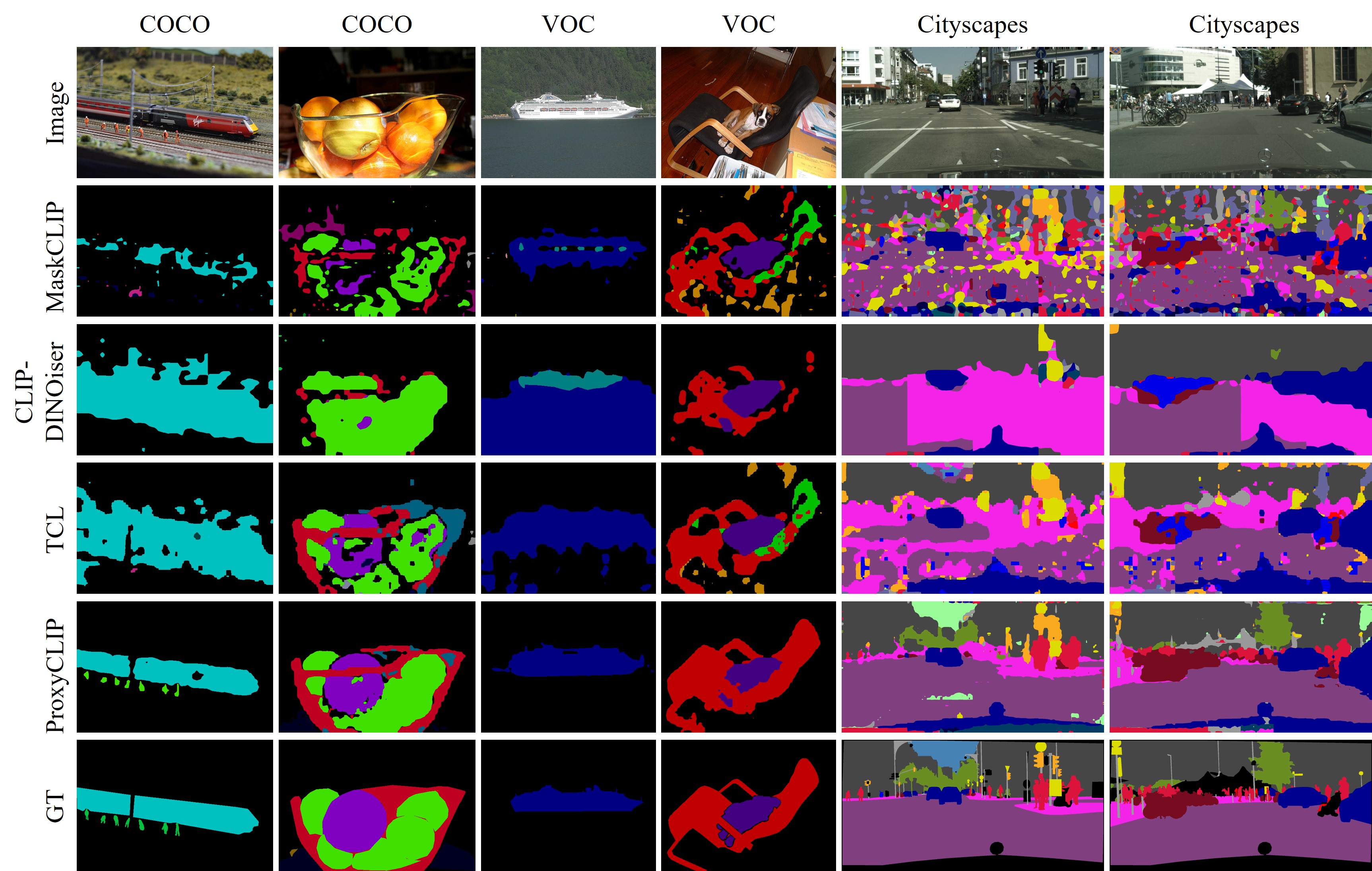}
    \caption{Qualitative comparison of semantic segmentation results.}
    \label{fig:visualization}
\end{figure}

\subsubsection{Qualitative results.}
\Cref{fig:visualization} demonstrates the qualitative comparison results of MaskCLIP \cite{zhou2022extract}, CLIP-DINOiser \cite{wysoczanska2023clip}, TCL \cite{cha2023learning} and our ProxyCLIP.
For a fair comparison, we do not use any post-refinement technique.
We observe that our proposed ProxyCLIP consistently produce higher quality and more precise segmentation maps compared to the other methods.
By leveraging the higher resolution and spatially coherent features of VFMs, ProxyCLIP successfully localizes smaller objects, such as the person in the first and the last two columns.
Moreover, ProxyCLIP produces image segments that accurately represent the scene layout (\eg, streetscape in the last two columns) and the boundaries of objects (\eg, the train in the first column and the ship in the third column), whereas other approaches tend to yield coarser results.

\subsection{Ablation study}

\begin{table}[!t]
  \caption{Comparison of open-vocabulary semantic segmentation performance under different models and architectures. ``Arch'' denotes the architecture, where ``B/16'' represents ViT-Base architecture with a patch size of 16.
  }
  \tabcolsep2.8pt
  \label{tab:VFMs}
  \centering
  \begin{tabular}{lc|ccccccccc}
    \toprule
    VFMs & Arch & VOC & Context & Object & VOC20 & Context59 & Stuff & City & ADE & Avg. \\
    \midrule
    \rowcolor{lightgray} \multicolumn{3}{l}{CLIP-ViT-B/16} &&&&&&&& \\
    MAE & B/16 & 52.2 & 30.4 & 30.8 & 76.3 & 33.5 & 23.1 & 30.1 & 17.1 & 36.7 \\
    SAM & B/16 & 59.3 & 33.6 & 35.4 & 80.4 & 37.0 & 25.0 & 37.0 & 19.1 & 40.8 \\
    SD & UNet & \textbf{60.3} & \textbf{34.7} & 36.5 & 82.0 & \textbf{38.2} & \textbf{25.9} & \textbf{37.7} & 19.6 & \textbf{41.8} \\
    DINOv2 & B/14 & 58.6 & 33.8 & \textbf{37.4} & \textbf{83.0} & 37.2 & 25.4 & 33.9 & \textbf{19.7} & 41.1 \\
    DINO & B/16 & 59.3 & 34.4 & 36.2 & 79.7 & 38.1 & 25.7 & 36.0 & 19.4 & 41.1 \\
    \midrule
    \rowcolor{lightgray} \multicolumn{3}{l}{OpenCLIP-ViT-H/14} &&&&&&&& \\
    MAE & B/16 & 54.7 & 29.8 & 32.2 & 80.6 & 32.9 & 21.8 & 34.9 & 19.4 & 38.3 \\
    SAM & B/16 & \textbf{63.5} & 34.1 & 36.7 & 84.0 & 37.9 & 25.0 & 41.1 & 22.0 & 43.1 \\
    SD & UNet & 62.8 & 34.6 & 34.7 & 84.9 & 38.5 & 26.2 & \textbf{41.5} & 22.9 & \textbf{43.3} \\
    DINOv2 & B/14 & 61.5 & 34.0 & 37.3 & \textbf{86.1} & 37.8 & 26.2 & 37.8 & 23.4 & 43.0 \\
    DINO & B/16 & 62.9 & \textbf{34.7} & \textbf{37.8} & 83.7 & \textbf{38.7} & \textbf{26.4} & 38.6 & \textbf{23.5} & \textbf{43.3} \\
  \bottomrule
  \end{tabular}
\end{table}
\begin{figure}[!t]
    \centering
    \includegraphics[width=1.0\linewidth]{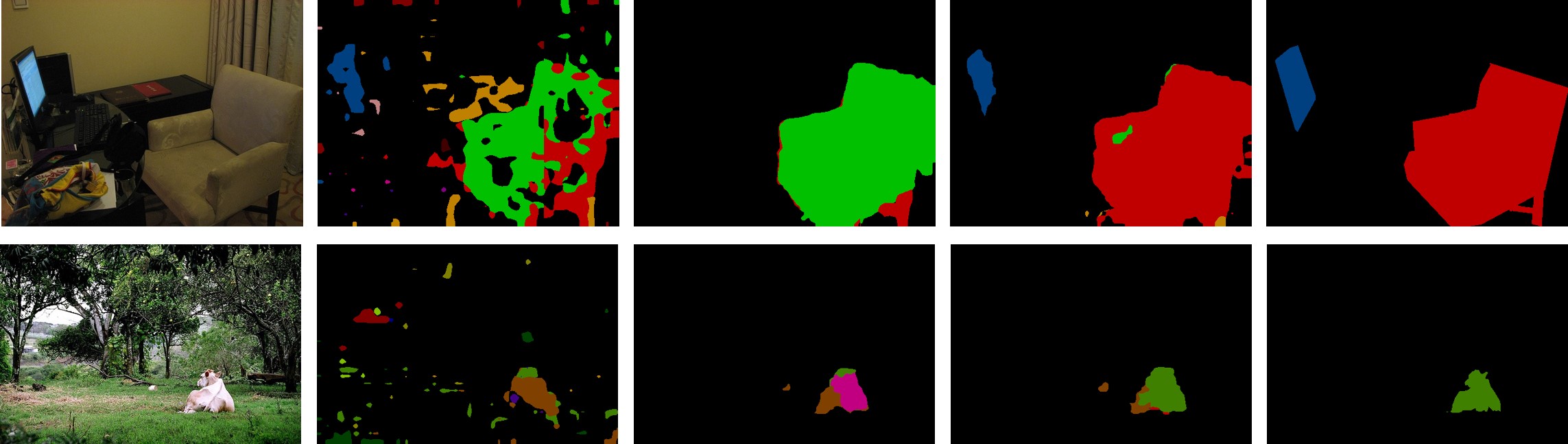}
    \tiny{
        \makebox[0.194\linewidth][c]{(1) Input}
        \makebox[0.194\linewidth][c]{(2) \makecell[c]{MaskCLIP \\ (CLIP-ViT-B/16)}}
        \makebox[0.194\linewidth][c]{(3) \makecell[c]{CLIP-ViT-B/16 \\ DINO-ViT-B/8}}
        \makebox[0.194\linewidth][c]{(4) \makecell[c]{OpenCLIP-ViT-H/14 \\ DINO-ViT-B/8}}
        \makebox[0.194\linewidth][c]{(5) Ground Truth}
    }
    \caption{Qualitative results on PASCAL VOC. (3)(4): Different configurations of ProxyCLIP. It is clear that proxy attention from DINO largely improves the spatial consistency while better CLIP backbone improves the semantics.}
    \label{fig:backbone}
\end{figure}
\subsubsection{Impact of the backbones.}
We then investigate the impact of various VFMs and CLIP backbones on the performance of ProxyCLIP.
In \cref{tab:VFMs}, we observe that the results of ProxyCLIP under the CLIP-ViT-B/16 using DINO \cite{caron2021emerging}, DINOv2 \cite{oquab2023dinov2, darcet2023vision}, and SAM \cite{kirillov2023segment} are very close, averaging 41.0 mIoU, as well as under the OpenCLIP-ViT-H/14. 
This indicates that these VFMs provide a comparable quality of proxy attention.
Interestingly, the self-supervised DINO performs as well as the supervised SAM. 
We attribute this to the low resolution of the input images (336$\times$336), which does not benefit SAM, as it relies on high-resolution inputs.
We also observe that, under CLIP-ViT-B/16 model, utilizing SD \cite{rombach2022high} as VFM yields better performance compared to using VFMs with ViT-B/16 or ViT-B/14 architecture.
However, SD features still lag behind DINO features with ViT-B/8 architecture (42.3 mIoU in \cref{tab:patch_size}).
On the other hand, we observe an improvement of around 2.0 mIoU for each variant when using OpenCLIP-ViT-H/14 models. 
This suggests that ProxyCLIP benefits from the enhanced semantic features of CLIP, confirming its effectiveness and generality.

We further conduct a qualitative comparison between different configurations of ProxyCLIP and MaskCLIP. 
To show the role of proxy attention, we further make qualitative comparison between different configurations of ProxyCLIP and MaskCLIP. \cref{fig:backbone}(2)$\rightarrow$(3) clearly shows that incorporating proxy attention from DINO-ViT-B/8 significantly enhances the local consistency of the segmentation maps. \cref{fig:backbone}(3)$\rightarrow$(4) highlights that utilizing a larger OpenCLIP backbone increases classification accuracy.

\begin{table}[!t]
  \caption{Comparison of open-vocabulary semantic segmentation performance under DINO architectures with different patch size.
  }
  \tabcolsep2.8pt
  \label{tab:patch_size}
  \centering
  \begin{tabular}{lc|ccccccccc}
    \toprule
    VFMs & Arch & VOC & Context & Object & VOC20 & Context59 & Stuff & City & ADE & Avg. \\
    \midrule
    \rowcolor{lightgray} \multicolumn{3}{l}{CLIP-ViT-B/16} &&&&&&&& \\
    DINO & S/16 & 59.1 & 34.2 & 35.8 & 79.8 & 37.9 & 25.6 & 35.8 & 19.2 & 40.9 \\
    \rowcolor{lightgreen} DINO & S/8 & 60.0 & 34.7 & 36.9 & 79.6 & 38.5 & 26.2 & 37.7 & 19.9 & 41.7 \\
    DINO & B/16 & 59.3 & 34.4 & 36.2 & 79.7 & 38.1 & 25.7 & 36.0 & 19.4 & 41.1 \\
    \rowcolor{lightgreen} DINO & B/8 & \textbf{61.3} & \textbf{35.3} & \textbf{37.5} & \textbf{80.3} & \textbf{39.1} & \textbf{26.5} & \textbf{38.1} & \textbf{20.2} & \textbf{42.3} \\
    \midrule
    \rowcolor{lightgray} \multicolumn{3}{l}{OpenCLIP-ViT-H/14} &&&&&&&& \\
    DINO & S/16 & 62.8 & 34.5 & 37.5 & 83.8 & 38.4 & 26.1 & 38.6 & 23.1 & 43.1 \\
    \rowcolor{lightgreen} DINO & S/8 & 63.5 & 34.8 & 38.0  & 82.2 & 38.9 & 26.3 & 41.7 & 23.9 & 43.7 \\
    DINO & B/16 & 62.9 & 34.7 & 37.8 & \textbf{83.7} & 38.7 & 26.4 & 38.6 & 23.5 & 43.3 \\
    \rowcolor{lightgreen} DINO & B/8 & \textbf{65.0} & \textbf{35.4} & \textbf{38.6} & 83.3 & \textbf{39.6} & \textbf{26.8} & \textbf{42.0} & \textbf{24.2} & \textbf{44.4} \\
  \bottomrule
  \end{tabular}
\end{table}

\begin{figure}[!t]
    \centering
    \includegraphics[width=0.9\linewidth]{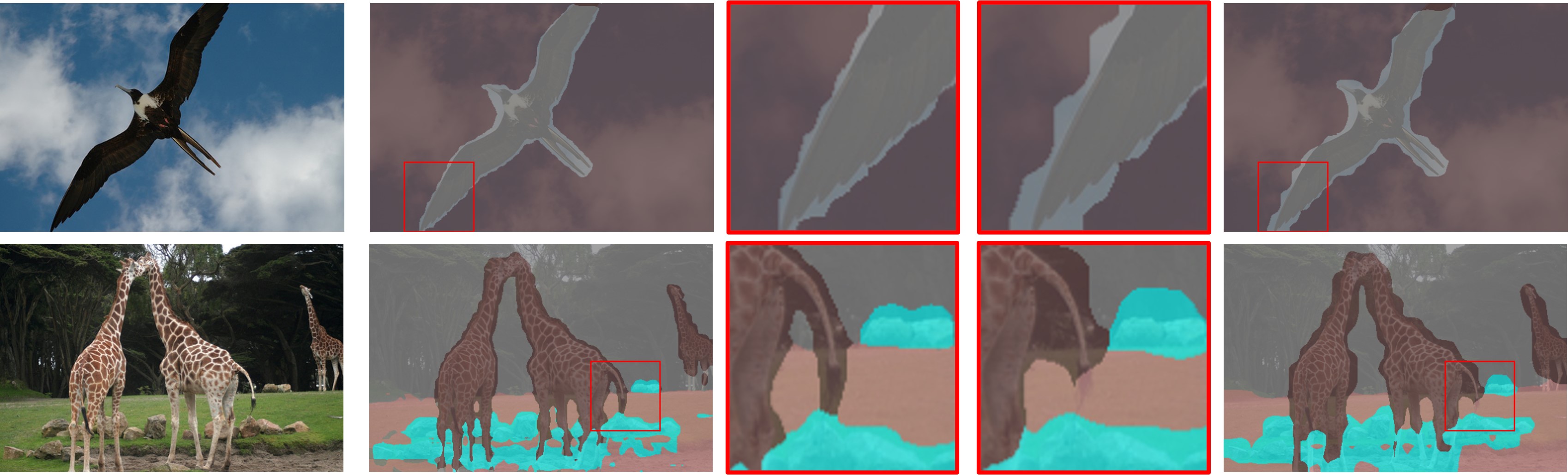}
    \makebox[0.22\linewidth]{\msmall{Input}}
    \makebox[0.35\linewidth]{\msmall{ProxyCLIP (DINO-ViT-B/8)}}
    \makebox[0.35\linewidth]{\msmall{ProxyCLIP (DINO-ViT-B/16)}}
    \caption{Qualitative comparison of different patch size. VFMs with smaller patch size of 8 helps ProxyCLIP to produce sharper boundaries.}
    \label{fig:patch_size}
\end{figure}

\subsubsection{Impact of the patch size.}
The resolution of model's feature maps plays a crucial role in determining the quality of the final segmentation results. 
Smaller patch sizes in the ViT architecture usually lead to higher-resolution feature maps, which in turn produces more detailed segmentation maps. 
In \cref{tab:patch_size}, our analysis of ProxyCLIP's performance with DINO at different patch sizes reveals that DINO with a smaller patch size (8) enables ProxyCLIP to achieve superior outcomes. 
Specifically, for the same model size, ProxyCLIP's performance with a patch size of 8 exceeds that of 16 by approximately 1.0 mIoU on average. 
This finding highlights ProxyCLIP's advantage when using VFMs with smaller patch sizes, yielding more precise and higher-quality segmentation maps. 
As depicted in \cref{fig:patch_size}, ProxyCLIP with DINO-ViT-B/8 produce sharper boundaries, such as the swings of the bird, and the tail of the giraffe. 

\subsubsection{Impact of the normalization and masking.}
In \cref{tab:norm_masking}, we evaluate the effect of our adaptive normalization and masking strategy on performance by using ProxyCLIP with CLIP-ViT-B/16 and DINO-ViT-B/8 architectures.
We observe that the combination of normalization and masking leads to the best results, increasing from an average of 25.0 mIoU to 42.3 mIoU. 
These results demonstrate that our proposed proxy attention mechanism with an adaptive normalization and masking strategy contributes to the success of incorporating VFMs and CLIP for open-vocabulary semantic segmentation.
Moreover, we present findings utilizing the attention scores from the last self-attention layer of each VFM. 
It is observed that employing the attention scores from DINO-ViT-B/8 and DINOv2-ViT-B/16 results in significantly poorer outcomes, while SAM-ViT-B/16 shows marginally lower performance (40.0 mIoU) compared to our proxy attention formula in \cref{eq:proxy_attention} (40.8 mIoU). 
This suggests that our proposed proxy attention method consistently delivers favorable outcomes across various VFMs.

\begin{table}[!t]
  \caption{Comparison of open-vocabulary semantic segmentation performance using different attention scores.
  }
  \tabcolsep2pt
  \label{tab:norm_masking}
  \centering
  \begin{tabular}{cc|ccccccccc}
    \toprule
    \multicolumn{2}{c|}{Attention} & VOC & Context & Object & VOC20 & Context59 & Stuff & City & ADE & Avg. \\
    \midrule
    \rowcolor{lightgray} Norm & Mask\\
    \ding{55} & \ding{55} & 32.5 & 18.8 & 17.0 & 70.0 & 22.1 & 14.0 & 15.8 & 10.3 & 25.0  \\
    \ding{51} & \ding{55} & 39.8 & 23.4 & 21.4 & 73.1 & 27.0 & 17.5 & 20.0 & 13.5 & 29.5 \\
    \rowcolor{lightgreen} \ding{51} & \ding{51} & \textbf{61.3} & \textbf{35.3} & \textbf{37.5} & 80.3 & \textbf{39.1} & \textbf{26.5} & \textbf{38.1} & \textbf{20.2} & \textbf{42.3} \\
    \midrule
    \multicolumn{2}{c|}{$\textup{Attn}_{\textup{DINO-B/8}}$} & 28.4 & 16.2 & 18.8 & 75.7 & 18.4 & 12.2 & 13.0 & 10.2 & 24.1 \\
    \multicolumn{2}{c|}{$\textup{Attn}_{\textup{DINOv2-B/14}}$} & 48.4 & 26.5 & 33.7 & \textbf{83.0} & 30.2 & 21.4 & 25.4 & 17.0 & 35.7 \\
    \multicolumn{2}{c|}{$\textup{Attn}_{\textup{SAM-B/16}}$} & 56.3 & 32.4 & 37.0 & \textbf{83.0} & 35.6 & 24.3 & 32.3 & 18.7 & 40.0 \\
  \bottomrule
  \end{tabular}
\end{table}

\section{Conclusion}

In this paper, we introduce ProxyCLIP, a training-free method using spatial feature correspondence from VFMs as proxy attention to enhance CLIP for open-vocabulary segmentation. 
Our comparative analysis of semantic coherence within CLIP's attention scores and VFMs' feature correspondences demonstrates that VFMs offer much better local correspondence than CLIP's inherent features, highlighting the benefits of integrating VFMs with CLIP for segmentation tasks. 
We explore various configurations of VFMs and CLIP and propose two main designs for the proxy attention: an adaptive normalization and masking strategy, and the use of small patch sizes. 
The experimental results indicate that ProxyCLIP not only significantly outperforms other training-free methods but is also superior to weakly-supervised approaches.

\noindent \textbf{Acknowledgments.} 
This study is supported under the RIE2020 Industry Alignment Fund – Industry Collaboration Projects (IAF-ICP) Funding Initiative, as well as cash and in-kind contribution from the industry partner(s).

%
%
\bibliographystyle{splncs04}
\bibliography{main}

\newpage
\appendix
\section*{Appendix}

\section{Implementation for Stable Diffusion.}
\label{sec:Different backbones}
Several recent studies \cite{xu2023open, li2023open, wang2023diffusion} have demonstrated the effectiveness of large-scale text-image diffusion models in open-vocabulary semantic segmentation tasks.
ODISE \cite{xu2023open} noted that the internal representations of stable diffusion models exhibit strong semantic coherence. 
Given the flexibility of our framework, our ProxyCLIP can also utilize stable diffusion models as VFM to extract dense visual representations for images.
Specifically, we employ the stable diffusion \cite{rombach2022high} model pre-trained on a subset of the LAION dataset as our VFM. 
We set the time step for the diffusion process to $t=0$ and extract feature maps from the 9-th block of UNet.
The input image is directly resized from 336$\times$336 to 672$\times$672.
Consequently, we obtain a feature map with dimensions $22 \times 22 \times 1280$, a downsampling factor of $15.3$.

\section{Hyperparameters.}
\begin{figure}[h]
  \centering
  \begin{subfigure}{0.4\linewidth}
    \includegraphics[width=1.0\linewidth]{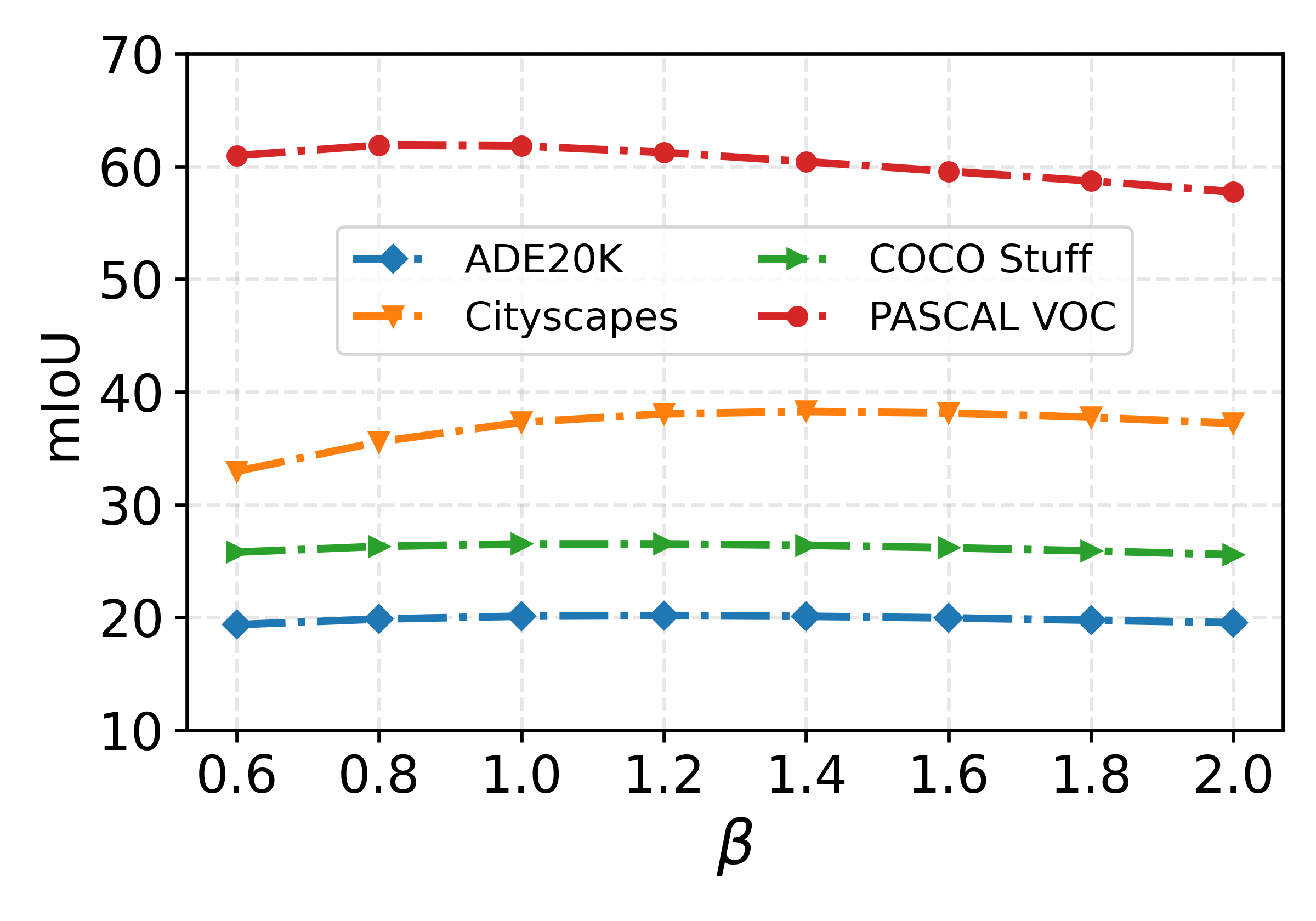}
    \label{fig:beta}
  \end{subfigure}
  \begin{subfigure}{0.4\linewidth}
  \includegraphics[width=1.0\linewidth]{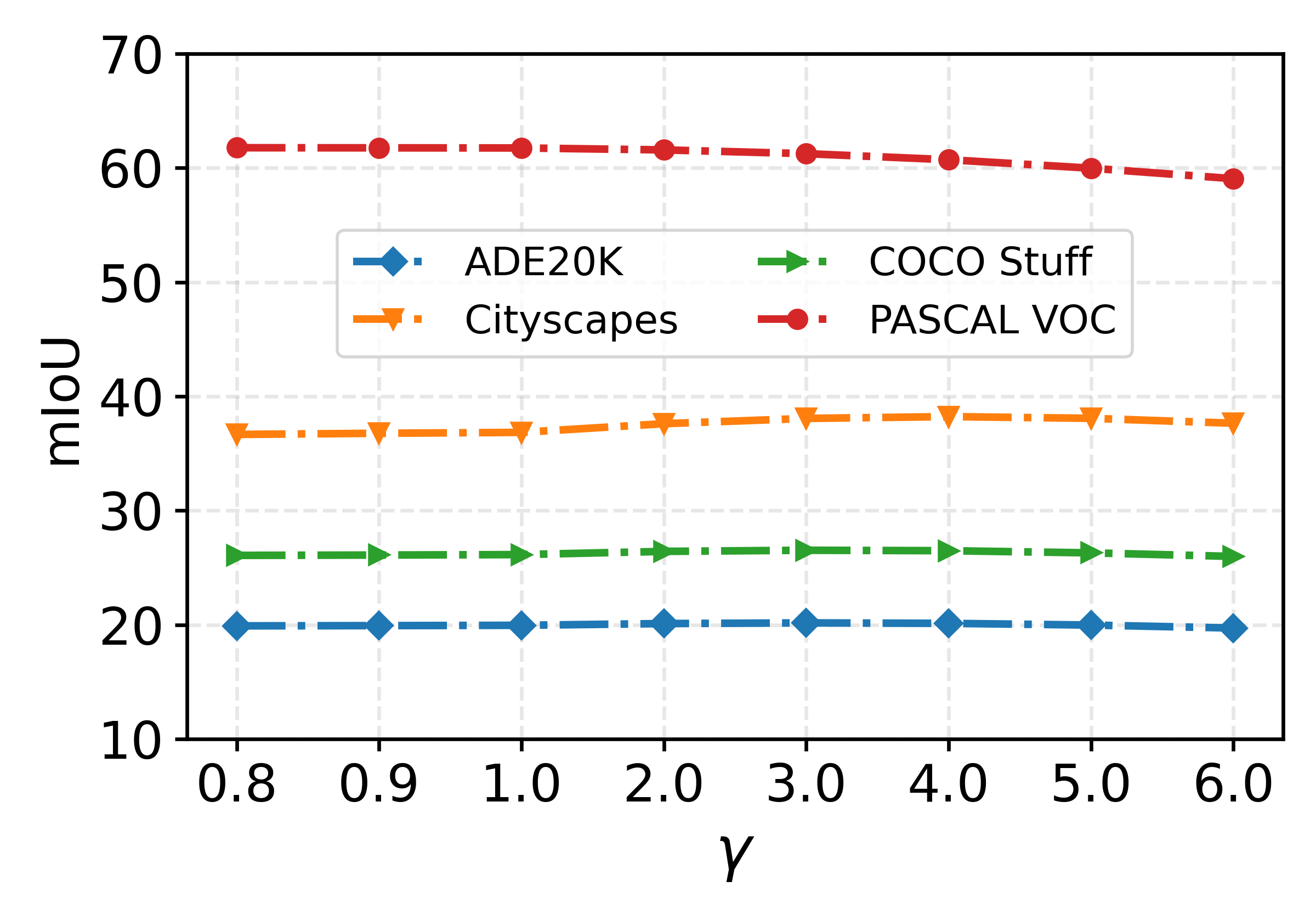}
    \label{fig:gamma}
  \end{subfigure}
  \caption{Open-vocabulary semantic segmentation results mIoU w.r.t. (left) the shifting factor $\beta$, and (right) the scaling factor $\gamma$.}
  \label{fig:parameters}
\end{figure}

We further conduct experiments to investigate the effects of varying the shifting and scaling factors in the normalization step.
As depicted in \cref{fig:parameters}, we examined the segmentation performance achieved across four datasets using different values for the shifting factor $\beta$ and the scaling factor $\gamma$.
Notably, we observe that ProxyCLIP consistently achieves good results when $\beta$ is localized within the range of 1.0 to 1.6 and $\gamma$ falls within the range of 2.0 to 5.0.
We set these two parameters to $\beta = 1.2$ and $\gamma = 3.0$ for all datasets by default.
These results further underscore the robustness of our normalization strategy within the proxy attention mechanism, affirming its efficacy in enhancing segmentation performance across diverse datasets.

\section{Efficiency comparison.}
We conduct an efficiency comparison among various training-free models during inference. 
These experiments are performed on an RTX 3090 GPU, using a batch size of 1 and an input image resolution of 336×336. 
Inference is conducted using half precision (fp16) for all models. 
The results in \cref{tab:Efficiency} indicate that ProxyCLIP, which utilizes an additional VFM (e.g., DINO-B/8), consumes more computational resources compared to models that use only the CLIP architecture.
However, adopting VFM with a smaller ViT architecture or a larger patch size may balance the accuracy and efficiency.
\begin{table}[h]
  \caption{Efficiency comparison of training-free methods based on CLIP-ViT-B/16 architecture. IPS: Image Per Second.}
  \tabcolsep5pt
  \renewcommand{\arraystretch}{1.0}
  \label{tab:Efficiency}
  \centering
  \begin{tabular}{c|ccccc}
    \toprule
     Models & \makecell[c]{Params $\downarrow$ \\ \small{\color{lightred}(M)}} & \makecell[c]{Speed $\uparrow$ \\ \small{\color{lightred}(IPS)}} & \makecell[c]{GPU $\downarrow$ \\ \small{\color{lightred}(MiB)}} & \makecell[c]{FLOPs $\downarrow$ \\ \small{\color{lightred}(G)}} \\
    \midrule
    CLIP & 142.7 & 72.5 & 3540 & 41.7 \\
    MaskCLIP & 142.7 & 74.1 & 3540 & 41.6\\
    GEM & 142.7 & 55.9 & 3716 & 51.5\\ \hline
    ProxyCLIP \scriptsize{(DINO-B/16)} & 224.5 & 52.9 & 3640 & 81.1\\
    ProxyCLIP \scriptsize{(DINO-B/8)} & 224.5 & 26.9 & 3926 & 253.3\\
  \bottomrule
  \end{tabular}
\end{table}

\section{Hard masking.}
To further verify the effectiveness of our adaptive normalization and masking strategy, we conduct an experiment focusing solely on the masking strategy. 
Implementing attention masking necessitates determining a suitable threshold, which proves challenging given the variability in median attention scores across different VFMs. 
To illustrate this, we perform experiments on the COCOStuff dataset using the masking strategy alone, with the threshold $\alpha$ varied within the range $[0.0, \dots, 0.8]$. 
The masking function is defined as follows:
\begin{equation}
    \mathcal{M}_{ij} = 
    \begin{cases}
        0, \ \ \ &A_{ij}\ge \alpha\\
        -\infty, \ \ \ &A_{ij}< \alpha
    \end{cases}
    \label{eq:masking_1}
\end{equation}
Results in \cref{tab:threshold} indicate that different VFMs may require different thresholds to achieve good results. 
In contrast, our adaptive normalization and masking strategy consistently delivers promising results across different VFMs.
\begin{table}[h]
  \caption{Comparison of using different thresholds on COCOStuff.
  }
  \tabcolsep8pt
  \renewcommand{\arraystretch}{1.0}
  \label{tab:threshold}
  \centering
  \begin{tabular}{c|ccccc|>{\columncolor{lightgreen}}c}
    \toprule
    $\alpha$ & 0.0 & 0.2 & 0.4 & 0.6 & 0.8 & Adaptive \\
    \midrule
    MAE & 15.2 & 16.9 & 19.8 & 23.0 & \colorbox{lightpink}{23.3} & 23.1\\
    SAM & 12.6 & 12.6 & 14.0 & 21.4 & \colorbox{lightpink}{25.2} & 25.0 \\
    DINOv2 & 15.5 & 22.4 & \colorbox{lightpink}{25.2} & 25.1 & 23.7 & 25.4\\
    DINO & 15.5 & 22.2 & \colorbox{lightpink}{25.8} & 24.4 & 22.0 & 26.5\\
  \bottomrule
  \end{tabular}
\end{table}

\section{Adopting attention embeddings of CLIP in PAM.}
To further evaluate our method, we conducte experiments using different attention embeddings from CLIP in the Proxy Attention Module (PAM). Specifically, we compare the use of $query$-$key$, $query$-$query$, and $key$-$key$ attention embeddings from CLIP with the use of $x$-$x$ features from Vision-and-Language Models (VFMs) in PAM (referred to as the Proxy).
The results, based on the CLIP-ViT-B/16 architecture, are summarized in \cref{tab:ablation}.
Notably, adopting $query$-$key$ embeddings of CLIP in PAM significantly enhances vanilla CLIP, improving from 11.7 mIoU to 26.1 mIoU. 
The performance is further improved to 38.2 when $query$-$query$ or $key$-$key$ embeddings are used.
Despite these enhancements, the performance of using CLIP embeddings in PAM is still inferior compared to the proposed proxy attention using VLM embeddings.

\begin{table}[h]
  \caption{Results of using CLIP's $q$ and $k$ embedings in PAM.
  }
  \tabcolsep3.5pt
  \renewcommand{\arraystretch}{1.0}
  \label{tab:ablation}
  \centering
  \begin{tabular}{c|ccccccccc}
    \toprule
    Attn & VOC & Context & Object & VOC20 & Context59 & Stuff & City & ADE & Avg. \\
    \midrule
    $q$-$k$ & 34.0 & 18.2 & 20.6 & 75.3 & 20.8 & 13.6 & 15.7 & 10.3 & 26.1 \\
    $q$-$q$ & 55.2 & 31.3 & 33.7 & 77.7 & 35.1 & 23.5 & 31.6 & 17.8 & 38.2 \\
    $k$-$k$ & 55.7 & 30.6 & 33.9 & 77.8 & 34.6 & 23.3 & 31.6 & 17.9 & 38.2 \\
    \rowcolor{lightgreen} Proxy & 61.3 & 35.3 & 37.5 & 80.3 & 39.1 & 26.5 & 38.1 & 20.2 & 42.3 \\
  \bottomrule
  \end{tabular}
\end{table}

\begin{table}[t]
  \caption{Comparison of open-vocabulary semantic segmentation performance under different models and architectures.
  }
  \tabcolsep2.8pt
  \label{tab:ADE847PC459}
  \centering
  \begin{tabular}{lcc|ccc}
    \toprule
    Method & &Annotation & ADE847 & PC459 &  Avg. \\
    \rowcolor{lightgray} \multicolumn{2}{l}{Fully-supervised}  &&&& \\
    ODISE \cite{xu2023open} &\textcolor{gray}{CVPR2023} & \ding{51} & 11.1 & 14.5 & 12.8 \\
    SAN \cite{xu2023side} &\textcolor{gray}{CVPR2023} & \ding{51} & 12.4 & 15.7 & 14.1\\
    X-Decoder \cite{zou2023generalized} &\textcolor{gray}{CVPR2023} &\ding{51} & 9.2 & 16.1 & 12.7\\
    OVSeg \cite{liang2023open} &\textcolor{gray}{CVPR2023} &\ding{51}& 9.0 & 12.4 & 10.7\\
    MaskCLIP \cite{ding2022open} &\textcolor{gray}{ICML2023} &\ding{51} & 8.2 & 10.0 & 9.1\\
    DeOP \cite{han2023open} &\textcolor{gray}{ICCV2023} &\ding{51} & 7.1 & 9.4 & 8.3\\
    MasQCLIP \cite{xu2023masqclip} &\textcolor{gray}{ICCV2023} & \ding{51} & 10.7 & 18.2 & 14.5\\
    GKC \cite{han2023global} & \textcolor{gray}{ICCV2023} & \ding{51} & 3.5 & 7.1 & 5.3\\
    MAFT \cite{han2023global} & \textcolor{gray}{NeurIPS2023} & \ding{51} & 12.1 & 15.7 & 13.9\\
    HIPIE \cite{wang2024hierarchical} & \textcolor{gray}{NeurIPS2023} & \ding{51} & 9.7 & 14.4 & 12.1\\
    \midrule
    \rowcolor{lightgray} \multicolumn{2}{l}{Weakly-supervised}  &&&& \\
    TCL \cite{cha2023learning} & \textcolor{gray}{CVPR2023} &\ding{55} & 4.9 & 5.3 & 5.1\\
    CLIP-DINOiser \cite{wysoczanska2023clip} & \textcolor{gray}{Arxiv2023} &\ding{55} & 7.1 & 8.4 & 7.8 \\
    \midrule
    \rowcolor{lightgray} \multicolumn{2}{l}{Training-free}  &&&& \\
    CLIP & \textcolor{gray}{ICML2021} &\ding{55} & 0.8 & 1.3 & 1.1\\
    MaskCLIP \cite{zhou2022extract} & \textcolor{gray}{ECCV2022} &\ding{55} & 3.6 & 4.6 & 4.1\\
    SCLIP \cite{wang2023sclip} & \textcolor{gray}{Arxiv2023} &\ding{55} & 4.9 & 6.3 & 5.6\\
    \rowcolor{lightgreen} ProxyCLIP & &\ding{55} & 11.1 & 9.9 & 10.5\\
  \bottomrule
  \end{tabular}
\end{table}

\begin{figure}[t]
  \centering
  \includegraphics[width=1.0\linewidth]{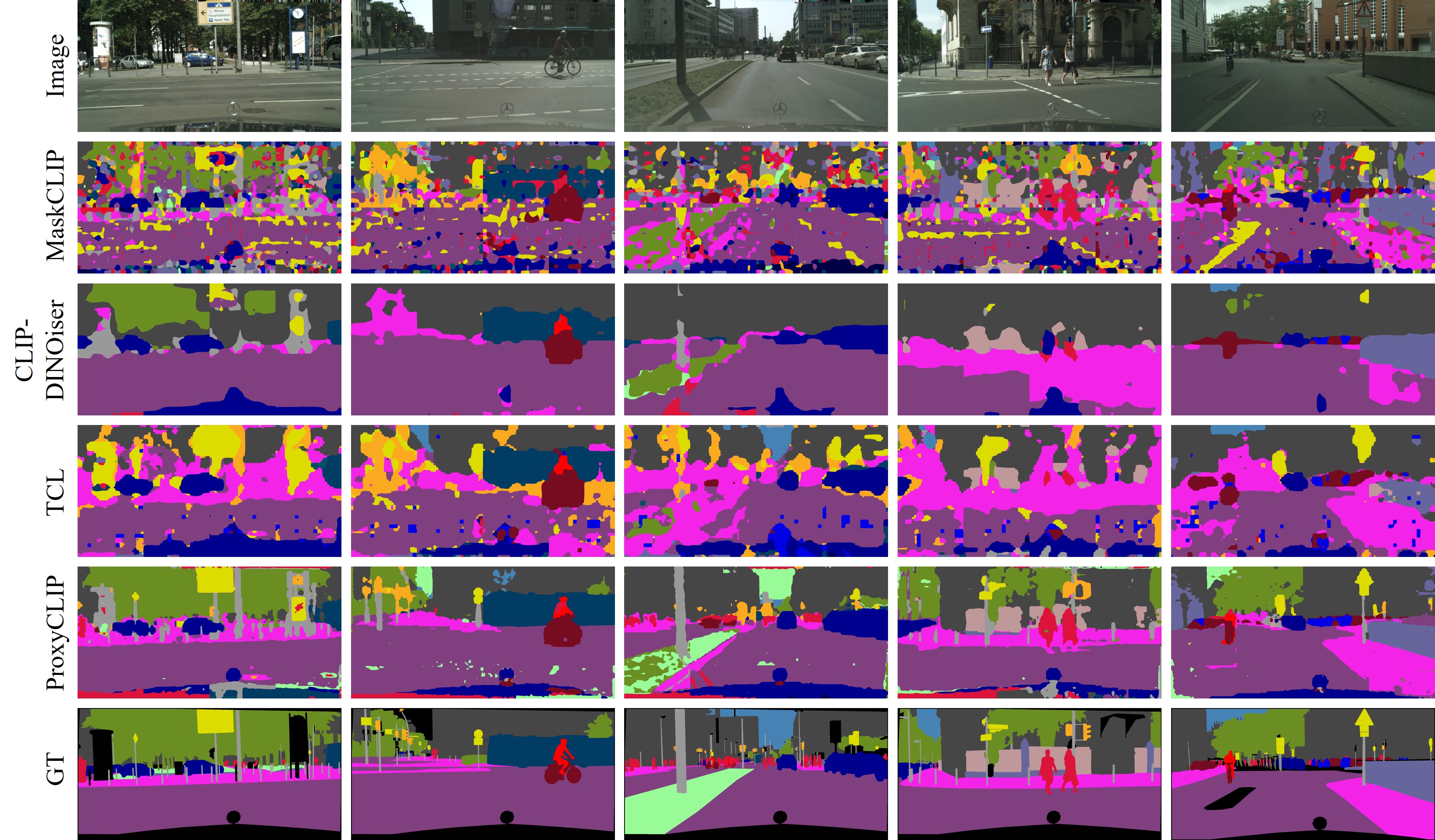}
  \caption{Additional qualitative comparison on the Cityscapes dataset.}
  \label{fig:vis_citys}
\end{figure}

\section{Additional quantitative results.}
\label{sec:Quantitative}
We further present a performance comparison on the ADE847 \cite{zhou2019semantic} and Pascal Context459 \cite{mottaghi2014role} (PC459) datasets, two challenging benchmarks containing 847 and 459 classes, respectively. 
These datasets are widely utilized for evaluating the performance of fully-supervised open-vocabulary semantic segmentation methods.
For fully-supervised open-vocabulary semantic segmentation methods, we directly cite the best results from their respective papers. 
For weakly-supervised methods, \ie, TCL and CLIP-DINOiser, we obtain results using the checkpoints provided by the authors. 
For training-free methods, we report results based on our implementation. 
To our knowledge, we are among the first to report results for weakly-supervised and training-free methods on these challenging datasets.

The experimental results are summarized in \cref{tab:ADE847PC459}. It's worth noting that fully-supervised methods typically achieve better results, as they undergo in-domain training using the COCOStuff training set with fully annotated labels.
Remarkably, ProxyCLIP, as a training-free method, achieves performance on par with fully-supervised methods. 
For instance, ProxyCLIP achieves 11.1 mIoU on the ADE847 dataset, surpassing the performance of most fully-supervised methods. 
Additionally, ProxyCLIP significantly outperforms all weakly-supervised and training-free methods, with an average mIoU of 10.5, compared to the 7.8 mIoU of CLIP-DINOiser.
We attribute this superiority to its proxy attention mechanism, which naturally inherits the robust local consistency of VFMs while maintaining CLIP's exceptional zero-shot recognition capacity.

\begin{figure}[!h]
  \centering
  \includegraphics[width=.95\linewidth]{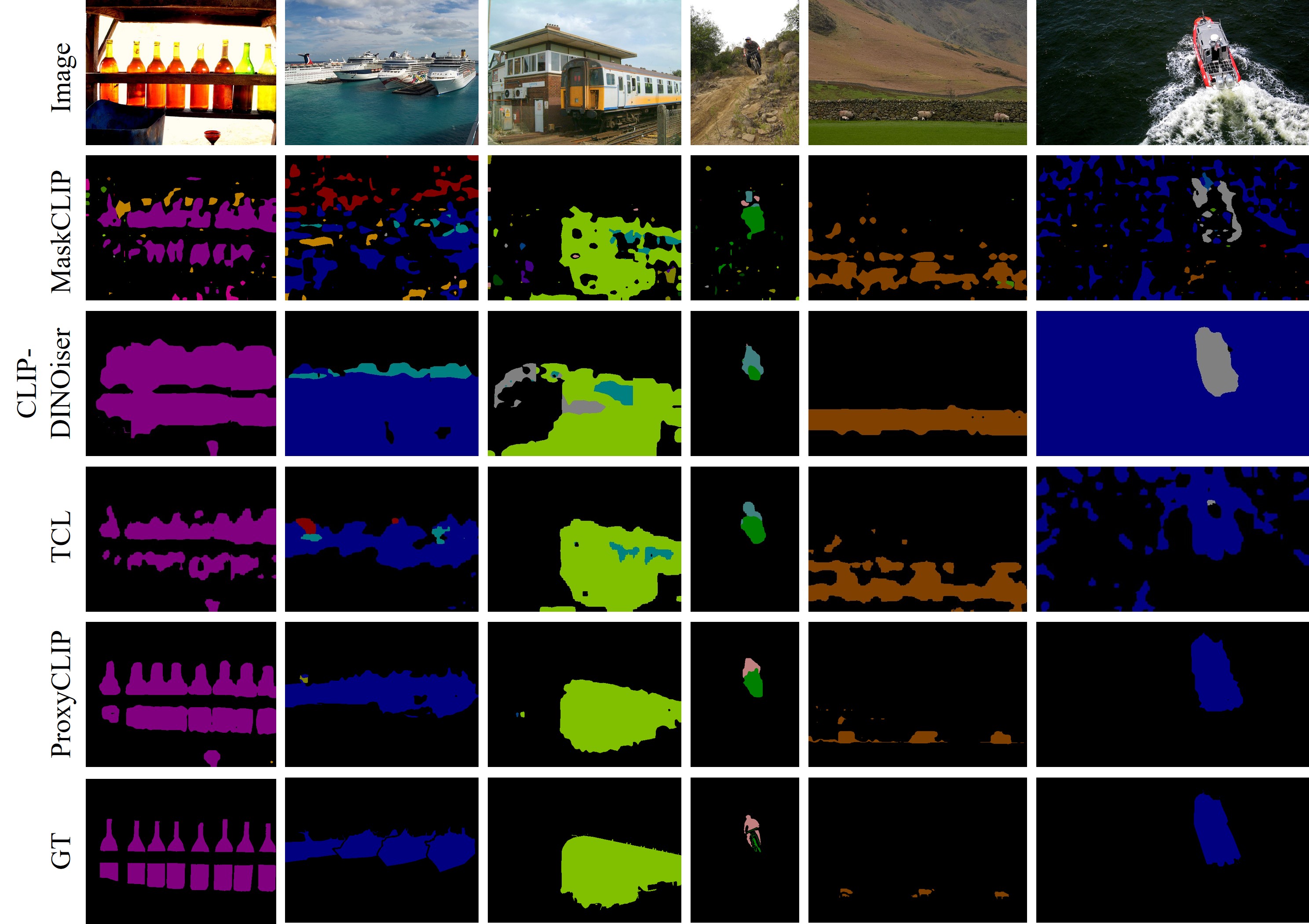}
  \caption{Additional qualitative comparison on the Pascal VOC dataset.}
  \label{fig:vis_voc21}
\end{figure}

\section{Additional qualitative results.}
\label{sec:Qualitative}
\begin{figure}[h]
  \centering
  \includegraphics[width=.95\linewidth]{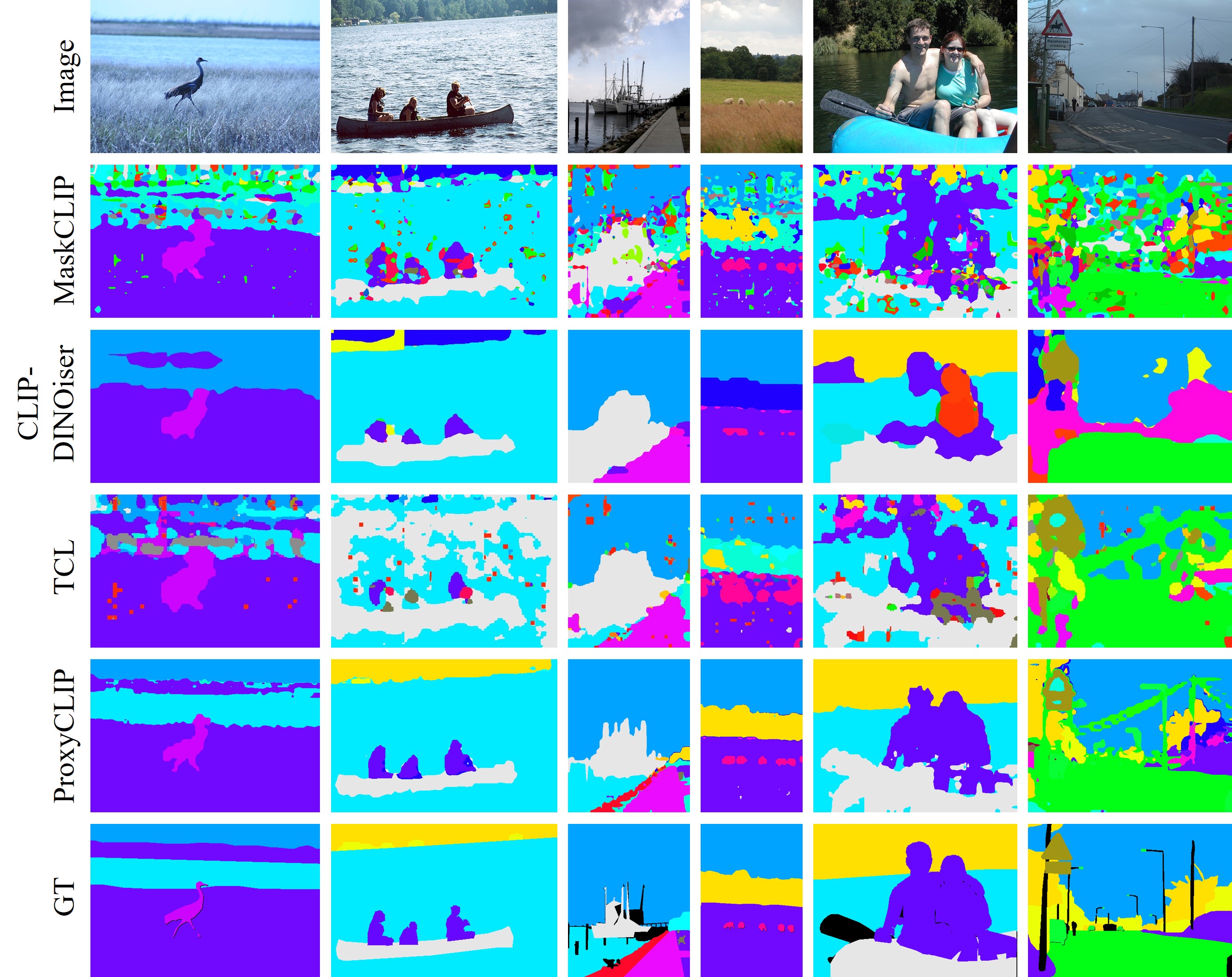}
  \caption{Additional qualitative comparison on the Pascal Context59 dataset.}
  \label{fig:vis_context59}
\end{figure}

We present additional qualitative results on three datasets: Cityscapes, Pascal VOC, and Pascal Context59, illustrated in \cref{fig:vis_citys,fig:vis_voc21,fig:vis_context59}, respectively.
In \cref{fig:vis_citys}, our ProxyCLIP demonstrates its ability to effectively segment regions belonging to different categories, including small objects.
\Cref{fig:vis_voc21,fig:vis_context59} showcase ProxyCLIP's capability to produce more accurate and high-quality segmentation maps with clearer boundaries for various objects.

\end{document}